# A General Algorithm for Approximate Inference and Its Application to Hybrid Bayes Nets

**Daphne Koller**     **Uri Lerner**     **Dragomir Angelov**

Computer Science Dept.
Stanford University
{koller,uri,drago}@cs.stanford.edu

## Abstract

The clique tree algorithm is the standard method for doing inference in Bayesian networks. It works by manipulating *clique potentials* — distributions over the variables in a clique. While this approach works well for many networks, it is limited by the need to maintain an exact representation of the clique potentials. This paper presents a new unified approach that combines approximate inference and the clique tree algorithm, thereby circumventing this limitation. Many known approximate inference algorithms can be viewed as instances of this approach. The algorithm essentially does clique tree propagation, using approximate inference to estimate the densities in each clique. In many settings, the computation of the approximate clique potential can be done easily using statistical importance sampling. Iterations are used to gradually improve the quality of the estimation.

## 1 Introduction

Bayesian networks (BNs) allow us to represent complex probabilistic models compactly and naturally. A range of inference algorithms, both exact and approximate, have been developed for the task of probabilistic reasoning using BNs — computing the probability of one or more events given some evidence. Until now, most of the Bayesian network models designed have admitted inference using one of the existing algorithms. However, over the last few years, there has been growing interest in extending the range of probabilistic reasoning applications [7, 8, 9]. Many of these domains present new challenges to BN technology: they are larger, more complex, and involve a wider range of probability models.

The most commonly used algorithm for BN inference is the *clique tree* algorithm [10, 16, 20]. The basic principle of the algorithm is to divide the set of variables into overlapping subsets called *cliques*. The inference is decomposed onto operations over the variables in a single clique. The results of a computation on one clique are transmitted to another, where they are in turn used in a subsequent computation. The complexity of the algorithm is determined by the size of the cliques; in discrete domains, it is roughly exponential in the number of variables in the largest clique.

The clique tree algorithm has two main advantages. First, it is an exact inference algorithm — it returns the correct answer, according to the joint distribution represented by the BN. Second, it can take advantage of structural properties of the domain — the conditional independences represented by the BN structure. These independences allow the construction of small cliques, reducing the complexity of the computation.

However, there are many situations where the clique tree algorithm breaks down. The main difficulty is that the intermediate results in the cliques can get too complex to be represented and manipulated effectively. Each intermediate result is a function from the space of values of its variables to $\mathbb{R}$. In discrete BNs, these functions are usually represented as tables, with an entry for each assignment in the appropriate domain. This representation is exponential in the number of variables in the expression. For some large BNs, e.g., QMR-DT [9], the cliques are too large to allow this exact representation of the factors. The problem is even more severe if we want to represent continuous or hybrid domains (ones involving both discrete and continuous variables). In this case, the intermediate results in the cliques are typically too complex to admit any closed form representation.

A completely different approach is taken by Monte Carlo inference algorithms such as Likelihood Weighting [19], Gibbs Sampling [17], some of which extend easily to hybrid BNs. The idea is to estimate the joint distribution as a set of (possibly weighted) samples. Unfortunately, the convergence of this algorithms can be quite slow in high-dimensional spaces, where the samples can explore only a small part of the space.

In this paper we present a general approach to combining exact and approximate inference, which in many cases achieves the best of both worlds. Like the clique tree algorithm, our approach builds a clique tree and propagates messages from one clique to another. However, rather than computing messages that correspond to exact intermediate results, our algorithm uses approximate inference techniques to compute and manipulate the messages.

This general scheme has many instantiations. These vary along two primary dimensions: the class of functions used to represent the intermediate results, and the approximation algorithm used to compute new intermediate results from previous ones. For example, in one instantiation, we might choose to represent intermediate results as mixtures of Gaussians. We can compute a new intermediate result, e.g., the product of two factors, by using random sampling to generate samples from the new distribution and then using statistical density estimation techniques (such as EM [4]) to estimate the new result.

As we will see, this scheme raises several interesting issues. For example, how do we best approximate the message that one clique sends to another? If we restrict the complexity of the message, then an approximation that tries to fit the entire function as well as it can may not be optimal. Rather, it may be better to provide a better approximation in those regions of the space that will turn out to be important. Of course, in early stages of the computation, we often do not know which parts of the space will later turn out to be important. Thus, an approximation that appears to be good now may turn out to be highly suboptimal later on. This observation leads us to an iterative approximation, where each intermediate result may be estimated several times.

We believe that our approach defines an interesting and useful class of approximate inference algorithms. Indeed, several well-known approximate inference algorithms can be viewed as special cases of this scheme. Our unified framework allows us to relate these algorithms to each other, to combine them, and even to improve their performance. More interestingly, we use this general framework to define a new algorithm for inference in hybrid Bayesian networks. We consider a particular instantiation of the general algorithm, which combines importance sampling with statistical density in order to compute the approximate messages and clique potentials. We present promising empirical results showing that our approach deals effectively with the difficult task of inference in such networks.

## 2 Preliminaries

In this section, we review two inference algorithms that will form the basis for our later discussion.

Assume that we have a given BN $\mathcal{B}$, over the variables $X_1, \ldots, X_n$. Each variable $X_i$ takes values in some domain $\text{Dom}(X_i)$, which may be discrete or continuous. As usual, the BN is a directed acyclic graph whose nodes are these variables. For a set of random variables $\mathbf{Y}$, we use $\text{Dom}(\mathbf{Y})$ to represent the joint domain of the variables in $\mathbf{Y}$. We use $\text{Parents}(X_i)$ to denote the parents of node $X_i$ in $\mathcal{B}$. Each node $X_i$ is also associated with a *conditional probability distribution (CPD)* $\phi_i$, which specifies the conditional distribution $P(X_i \mid \text{Parents}(X_i))$, i.e., for every value $\mathbf{y} \in \text{Dom}(\text{Parents}(X_i))$, it defines a probability distribution (or density function) over $\text{Dom}(X_i)$. We use $\phi_i$ to denote the function from $\text{Dom}(X_i \cup \text{Parents}(X_i))$ to $I\!R$. In discrete domains, a CPD is often implemented as a table, which simply lists the appropriate probability for each instantiation in $\text{Dom}(X_i \cup \text{Parents}(X_i))$.

The clique tree algorithm [16, 20] is the algorithm most commonly used to perform exact inference in discrete Bayes Nets. The basic data structure used by the algorithm is called a *clique tree* or a *join tree*. The clique tree is an undirected tree whose nodes $\mathbf{C}_1, \ldots, \mathbf{C}_k$ are called *cliques* or *clusters*. Each clique $\mathbf{C}_i$ is defined to contain some subset of the random variables in the BN. Each CPD $\phi_i$ in $\mathcal{B}$ must be assigned to some clique, which must contain $X_i$ and $\text{Parents}(X_i)$. We use $\mathcal{E}$ to denote the set of edges in the clique tree. Each edge is associated with a *sepset*, which is the set of random variables in the intersection of the two cliques at the endpoints of the edge. We use $\mathbf{S}_{i,j}$ to denote the sepset between $\mathbf{C}_i$ and $\mathbf{C}_j$, and $\mathbf{C}_{i \setminus j}$ to denote $\mathbf{C}_i - \mathbf{S}_{i,j}$.

There are several ways of executing inference over this type of data structure. We focus on the *Shafer-Shenoy* algorithm [20], which is the most suitable for our purposes, as it avoids division operations. Initially, each $\phi_i$ is assigned to some clique $\mathbf{C}_j$; we can only assign this CPD to a clique that contains $X_i \cup \text{Parents}(X_i)$. We use $\Phi(\mathbf{C}_j)$ to denote the set of CPDs assigned to $\mathbf{C}_j$. Each clique executes local computations over the variables in the clique, which correspond to multiplying messages and CPDs, and marginalizing the result over some of the variables in the clique (by summing out the rest). The results of these computations are sent as messages to neighboring cliques, which, in turn, use them in their own computation. A clique $\mathbf{C}_i$ is ready to send a message $\eta_{i \to j}$ to its neighbor $\mathbf{C}_j$ when it has received messages from all of its other neighbors. The algorithm terminates when each clique $\mathbf{C}_i$ has sent messages to all of its neighbors. For simplicity, we model the algorithm as incorporating evidence by multiplying the CPDs in each clique with an indicator function $\varepsilon_i$ for all the evidence relevant to that clique. We use $\phi_i^*$ as shorthand for $\prod_{\phi_k \in \Phi(\mathbf{C}_i)} \phi_k \times \varepsilon_i$.

---

**Algorithm 1: Exact clique tree propagation**
01: **for** each $\mathbf{C}_i$ and each $\mathbf{c} \in \text{Dom}(\mathbf{C}_i)$
02: **repeat**
03:    choose $\mathbf{C}_i$ that received all incoming msgs
          except (perhaps) $\eta_{j \to i}$
04:    compute $\tau_{i \to j} = \prod_{(j', i) \in \mathcal{E}, j' \neq j} \eta_{j' \to i} \times \phi_i^*$
05:    compute $\eta_{i \to j} = \sum_{\text{Dom}(\mathbf{C}_{i \setminus j})} \tau_{i \to j}$

06: **until** all messages sent

07: compute $\psi_i = \prod_{(j,i) \in \mathcal{E}} \eta_{j \rightarrow i} \times \phi_i^*$

08: normalize each $\psi_i$

We call the functions $\tau_{i \rightarrow j}$ *intermediate factors*. The final result of the computation $\psi_i$ is a *clique potential* at clique $i$. The clique potential $\psi_i$ represents the posterior distribution over the variables $\mathbf{C}_i$ conditioned on the evidence.

At a high level, this algorithm extends trivially to the case of continuous variables: we simply replace the summation in line 5 with an integration.

In addition to exact inference, Bayesian networks also support a range of approximate inference algorithms. Of these, the most commonly used approach is *Monte Carlo* sampling. Sampling is a general-purpose technique for inference in probabilistic models; its applicability goes far beyond Bayesian networks. The basic idea is that we can approximate the distribution by generating independent samples from it. We can then estimate the value of a quantity relative to the original distribution by computing its value relative to our samples.

Let $p$ be a probability distribution over some space $\Omega$. If $\Omega$ is discrete, the expectation of any function $f : \Omega \mapsto \mathbb{R}$ is defined as $E_p(f) = \sum_{\omega \in \Omega} f(\omega) p(\omega)$. (In the continuous case, we simply replace the summation by an integral.) Now, assume that we can generate a set of random samples from $p$, $\omega[1], \ldots, \omega[M]$. We can view these samples as a compact approximation to the distribution $p$. Therefore, we can approximate $E_p(f) \approx \frac{1}{M} \sum_{j=1}^{M} f(\omega[j])$. The main problem is that our distribution $p$ is often not one from which we can sample directly. For example, in a Bayesian network, we can easily generate samples from the prior distribution. However, it is far from trivial to generate samples from a posterior distribution. *Importance sampling* provides one approach for dealing with this difficulty.

Assume that our desired sampling distribution $p$ is given as some probability function $p'$, which may not even be normalized. We cannot sample from $p'$. However, we have some probability distribution $q$ from which we can sample. The only requirement is that $q(\omega) > 0$ whenever $p'(\omega) > 0$. We will approximate $p$, the normalized version of $p'$, by sampling from $q$. More precisely, we will generate samples $\omega[1], \ldots, \omega[M]$ from $q$. To each $\omega[j]$ we will assign a *weight* $w[j] = p'(\omega[j])/q(\omega[j])$. If we normalize the weights of our $M$ samples to sum to 1, we can view this set of weighted samples as an approximation to $p$. It can easily be shown [17] that $E_p[f] \approx \sum_{m=1}^{M} f(\omega[m]) \frac{w[m]}{\sum_j w[j]}$.

## 3   The General Inference Algorithm

Our approach is based on the following simple idea, used in several papers (see Section 6). Rather than requiring that the algorithm produce the exact result when computing fac-tors (in lines 4, 5, and 7 of Alg. 1), we allow it to produce an approximation to it — one that admits a compact representation. Of course, this general idea can be implemented in many ways. We can choose a variety of representations for the factors. Clearly, the choice of representation has a strong effect on the quality of our approximation. We need also determine how to implement the various operations over factors specified in the algorithm. For example, in lines 4 and 5, our approximate inference algorithm will need to take one set of (approximate or exact) messages, multiply them by the CPDs and the evidence function, and generate an approximation to the outgoing message $\eta_{i \rightarrow j}$.

One issue that complicates both the representation and the approximation is that neither the intermediate factors nor the messages are guaranteed to be probability densities. For example, in some circumstances a message may be simply a CPD. As the parents do not have distributions, the CPD is not a joint distribution over the variables it mentions. As density estimation is much more circumscribed than general-purpose function approximation, we choose to restrict attention to densities. As in [14], we assume that the ranges of all continuous variables are bounded. We normalize each factor which is not a density by ascribing uniform distributions to variables whose distribution is not defined in the factor. Note that a multiplication by a constant factor does not influence the correctness of the algorithm, as the clique potentials are normalized at the end.

We can now utilize any of a variety of density estimators for representing our factors. The available approximation techniques will, of course, depend on the choice of representation. If, for example, we were to choose a single Gaussian as the representation for factors, we could compute the optimal Gaussian approximation relatively easily, simply by estimating the first and second moments of the distribution represented by our factor. A more general approach that we can use for approximation combines (importance) sampling and statistical density estimation. Consider the task of computing the final clique potential. We want to generate samples from this distribution, and then apply a density estimation algorithm to learn the associated density. In certain (rare) cases, we can sample directly from the resulting density; more often, we cannot. In these cases, importance sampling provides a solution, as described in the previous section.

Let us assume that we have chosen some representation for the factors. How do we best approximate $\eta_{i \rightarrow j}$? The obvious solution is to try and find the approximation $\hat{\eta}_{i \rightarrow j}$ that minimizes some metric (a reasonable choice would be the KL-distance [2]) between the "correct" message $\eta_{i \rightarrow j}$ (normalized to produce a density) and the approximation $\hat{\eta}_{i \rightarrow j}$. Unfortunately, this intuition is flawed. We do not need to approximate the message per se. What we really want is to create the message that will give the best result when we approximate the potential $\psi_j$. This goal might lead to very different requirements.

Intuitively, this phenomenon is clear. A good approximation relative to KL-distance will be more accurate in the regions where the density has high value. If the message is then multiplied with additional factors in $C_j$, leading to a very different function, regions that had very low weight in the original message can now have high weight. The approximation may now be a very poor one. In general, even if an approximation to a message has low KL-distance, it does not imply that it the result of multiplying that message with other factors will also be a good approximation.

We now provide a rough analysis of this phenomenon. We stress that this analysis is not intended to be an exact bound, but rather to give us insight on the design of the algorithm. The computation of the true potential is described in line 7 of the algorithm. It is easy to verify that $\psi_j$ is proportional to a product of $\eta_{i \to j}$ and several other factors (messages, CPDs, and evidence term). We define $rest_j$ to be the products of these other factors (actually, $rest_j$ is simply $\tau_{j \to i}$). Thus, rather than minimizing $D(\eta_{i \to j} \| \hat{\eta}_{i \to j})$, we should strive to reduce $D(\alpha_1 \eta_{i \to j} rest_j \| \alpha_2 \hat{\eta}_{i \to j} rest_j)$, where $\alpha_i$ are normalizing constants.

Note that $\alpha_1 \eta_{i \to j} rest_j$ is the best approximation to the true potential we can achieve in clique $Y$ given the other (possibly approximate) incoming messages. Hence, we want to minimize $D(\alpha_1 \eta_{i \to j} rest_j \| \alpha_2 \hat{\eta}_{i \to j} rest_j)$, which is equal to $\sum_{s \in S_{i,j}} \hat{\psi}_j(s)(\alpha_1 \eta_{i \to j}(s))/(\alpha_2 \hat{\eta}_{i \to j}(s))$. Hence, the penalty we get for inaccuracies in $\hat{\eta}_{i \to j}$ is proportional to $\hat{\psi}_j(s)$ — the marginal density of the new potential in $C_j$ over the sepset's variables $S_{i,j}$. Hence, $\hat{\eta}_{i \to j}$ should be accurate in areas where $\hat{\psi}_j$ is big and not necessarily in areas where $\eta_{i \to j}$ is big. (A similar derivation appears in [14].)

The problem of finding an approximation to a distribution $q$ whose accuracy is optimal relative to some other distribution $p$ is a nontrivial one. Most density estimation algorithms, if applied naively to the task of learning $q$, would construct an approximation that was more accurate in regions where $q$ had high value. For example, if we generated samples from $q$, we would end up with many more samples, and thereby a much more refined estimate of the density, in regions where $q$ has high mass. We have explored several different heuristics for this task, all of which follow roughly the same scheme: generate a density estimate for $p$, and then modify it to fit $q$. The approach which worked best in our experiments is quite simple. We generate samples from $p$ (which in our case is $\hat{\psi}_j(S_{i,j})$), thereby generating more samples in the "right" regions. We then weight the samples by $q/p$, making them estimates to the desired distribution. Finally, we apply density estimation to the resulting weighted samples.

Our solution above estimated $\eta_{i \to j}$ using the approximate potential $\hat{\psi}_j$ for $C_j$. However, when we are computing

$\eta_{i \to j}$, we do not yet have $\hat{\psi}_j$; indeed, we are computing $\eta_{i \to j}$ for the express purpose of using it to compute $\hat{\psi}_j$. There are two ways to resolve this problem. The first is to use our current best approximation to the potential in $C_j$. While this approach might seem plausible, it is misguided. We should be sending a message from $C_i$ to $C_j$ only if $C_i$ has information not available in $C_j$. Thus, any approximation we might have in $C_j$ is known to be flawed, and we should not use it to guide our message approximation. The second solution is based on the observation that we are only interested in the marginal of $\hat{\psi}_j$ over $S_{i,j}$. We have another approximation to this marginal — $\hat{\psi}_i(S_{i,j})$. As we mentioned, the fact that we need to send a message from $C_i$ to $C_j$ indicates that $C_i$ is more informed (in some sense) than $C_j$. Thus, we use $\hat{\psi}_i(S_{i,j})$ to guide our approximation to the message $\hat{\eta}_{i \to j}$. Of course, this intuition is only partially valid: there may be other ways in which $C_j$ is more informed than $C_i$. In general, we note that the choice of function to guide our approximation is only a heuristic, and does not influence the "correctness" of the message. This function is used only to derive the sampling distribution; the weights on the samples guarantee that the actual function being approximated is $\hat{\eta}_{i \to j}$.

This observation immediately leads to the final major improvement to our basic algorithm: the iterative improvement of the messages. When we send a message from $C_i$ to $C_j$, we rely on our current approximation $\hat{\psi}_i$. Later on, $C_i$ may get a message from $C_j$ or from some other clique, which leads to a more accurate approximation $\hat{\psi}_i$. If $C_i$ sends a message to $C_j$ using this new improved approximation as the basis for approximating $\eta_{i \to j}$, the resulting approximate message $\hat{\eta}_{i \to j}$ will almost certainly be more reliable. If $\hat{\eta}_{i \to j}$ is more reliable, then outgoing messages from $C_j$ will also be more reliable. Thus, our algorithm is an iterative improvement algorithm. It continues to improve its current best estimate for each clique potential $\hat{\psi}_i$ to produce more accurate message. These more accurate messages are, in turn, used to compute more accurate approximate potentials in the destination cliques.

We are now ready to present our general algorithm. At this point we leave many issues open to different implementations. We later show how several existing algorithms are instances of this general scheme, and present one new instantiation of it in detail.

The first phase is virtually the same as Algorithm 1, except that an approximate computation is used in place of an exact one. We omit details. In the second phase, the algorithm iterates over the cliques, trying to improve its approximation of the potentials and the messages:

**Algorithm 2: Iteration phase for approximate propagation**
01: **repeat**
02:   choose some clique $C_i$
03:   approximate $\hat{\psi}_i' = \prod_{(j,i) \in \mathcal{E}} \hat{\eta}_{j \to i} \times \phi_i^*$

```
04:    set $\hat{\psi}_i = \hat{\psi}'_i$
05:    for some or all $(i, j) \in \mathcal{E}$
06:        generate an estimate $\mu_{i \to j}$ to $\sum_{\text{Dom}(\mathbf{C}_{i \setminus j})} \hat{\psi}'_i$
07:        reweight $\mu_{i \to j}$ by $(\prod_{j' \neq j} \hat{\eta}_{j' \to i} \times \phi_i^*)/\hat{\psi}_i$
08:    until convergence
```

Intuitively, the algorithm repeatedly selects a clique for improvement (using some criterion). It estimates the revised clique potential, and uses that as the basis for re-estimating some or all of the messages. As described above, each message is first estimated to match the current potential. It is then reweighted to make it represent the correct function that $\mathbf{C}_i$ should send to $\mathbf{C}_j$.

Clearly, there are many instantiations to this general schema. For example, the algorithm can select which clique potential to update in a variety of ways. The simplest and cheapest approach, which is the one we used, is to execute repeated upward and downward passes, mirroring the computation in standard clique tree inference. Also, as we mentioned earlier, the approximation phase can be done in many ways. One approach is to use importance sampling, possibly followed by density estimation. In our particular instantiation of the algorithm, all approximations were done using this technique. In the initial calibration phase, we use an approximation to the prior distribution (before any evidence is inserted) as the sampling distribution. We obtain this approximation to the prior distribution by sampling the original BN, using the simple top-down sampling algorithm. Our ability to do this effectively is due precisely to the fact that we are sampling from the prior. In the iteration phase, we use our current posterior $\hat{\psi}_i$ as the sampling distribution, as discussed above.

## 4   Algorithm for hybrid BNs

In this section, we describe in detail one instantiation of the general approach described above, designed to perform effective inference in general hybrid BNs.

Algorithms for exact inference in hybrid BNs are applicable only to very narrow families of BNs (multivariate Gaussians [18] or, in some sense, conditional Gaussians [15]). Our algorithm is very general in the sense it can deal with virtually any function for the CPD, as long as we can compute the value of the function at a point, and sample a value for the node from the conditional density defined by a particular assignment to its parents. This flexibility allows our algorithm to deal with significantly more interesting models. In particular, our algorithm is capable of handling arbitrary dependencies, including ones where a discrete node has continuous parents. Our initial implementation incorporates a few basic yet flexible CPDs that allow us to represent a fairly wide range of models.

For a discrete child with discrete parents, we implemented the standard conditional probability table model. For a continuous node with discrete and continuous parents, we implemented the standard conditional linear Gaussian model [15]. In this model the child's mean is a linear function of the continuous parents, and its covariance is fixed. We have a separate linear Gaussian model for the child for every value of the discrete parents. We also allow uniform distributions.

While these CPDs are fairly standard, we also defined a new class of CPDs for modeling the dependence of a discrete node on discrete and continuous parents. This CPD is an interesting and useful generalization of the standard *softmax* density [1]. As for the conditional linear Gaussian CPD, our softmax CPD will have a separate component for each instantiation of the discrete parents. Therefore, it suffices to describe the model for the case where the discrete node has only continuous parents. Intuitively, the softmax CPD defines a set of $R$ regions (for some parameter $R$ of our choice). The regions are defined by a set of $R$ linear functions over the continuous variables. A region is characterized as that part of the space where one particular linear function is higher than all the others. Each region is also associated with some distribution over the values of the discrete child; this distribution is the one used for the variable within this region. The actual CPD is a continuous version of this region-based idea, allowing for smooth transitions between the distributions in neighboring regions of the space.

More precisely, let $C$ be a discrete variable, with continuous parents $\mathbf{Z} = \{Z_1, \ldots, Z_m\}$. Assume that $C$ has $k$ possible values, $\{c_1, c_2, \ldots, c_k\}$. Each of the $R$ regions is defined via two vectors of parameters $\boldsymbol{\alpha}^r, p^r$. The vector $\boldsymbol{\alpha}^r$ is a vector of weights $\alpha_0^r, \alpha_1^r, \ldots, \alpha_m^r$ specifying the linear function associated with the region. The vector $p^i = \{p_1^r, \ldots, p_k^r\}$ is the probability distribution over $c_1, \ldots, c_k$ associated with the region (i.e., $\sum_{j=1}^k p_j^r = 1$). The CPD is now defined as: $P(C = c_j \mid \mathbf{Z}) = \sum_{r=1}^R w^r p_j^r$ where $w^r = \frac{\exp(\alpha_0^r + \sum_{i=1}^m \alpha_i^r Z_i)}{\sum_{q=1}^R \exp(\alpha_0^q + \sum_{i=1}^m \alpha_i^q Z_i)}$. In other words, the distribution is a weighted average of the region distributions, where the weight of each "region" depends exponentially on how high the value of its defining linear function is, relative to the rest.

The power to choose the number of regions $R$ to be as large as we wish is the key to the rich expressive power of the generalized softmax CPD. Figure 1 demonstrates this expressivity. In Figure 1(a), we present an example CPD for a binary variable with $R = 4$ regions. In Figure 1(b), we show how this CPD can be used to represent a simple classifier. Here, $C$ is a sensor with three values: low, medium and high. The probability of each of these values depends on the value of the continuous parent $X$. Note that we can easily accomodate a variety of noise models for the sensor: we can make it less reliable in borderline situations by making the transitions between regions more moderate; we can make it inherently more noisy by having the probabilities of the different values in each of the regions be farther away from 0 and 1.

Perhaps the most important decision in using our general algorithm is the decision on the actual representation of the potentials. In our case we needed a representation that enables us to express hybrid density functions. Furthermore, we were looking for a representation from which samples can be generated easily, and which can, in turn, be easily estimated from a set of samples.

We chose to focus on *Density Trees* [13]. The structure of a density tree resembles that of a classification decision tree; however, rather than representing the conditional distribution of a class variable given a set of features, a density tree simply represents an unconditional density function over its set of random variables $\mathbf{X}$. A density tree has two types of nodes: interior nodes and leaves. An interior node $u$ defines some set of mutually exclusive and exhaustive events $E_1^u, \ldots, E_k^u$. Each branch from $u$ corresponds to one of the events. The definition allows for arbitrary events; for example, we may have an interior node with three branches, corresponding to $X_2 < 3$, $3 \leq X_2 < 5$ and $X_2 \geq 5$. This definition generalizes that of [13]. The edge corresponding to the outcome $E_i^u$ is labeled with the conditional probability of $E_i^u$ given all the events on the path from the root to node $u$. We define the probability of a path as the product of all the probabilities along the path, i.e., the probability of the conjunction of the events along the path.

A leaf in the tree corresponds to some probability distribution consistent with events on the path leading to it (i.e., everything that is inconsistent with the path events has probability zero). To find the probability of some instantiation $\mathbf{x}$, we traverse the tree from the root to a leaf on the unique path which is consistent with $\mathbf{x}$. During this traversal, we compute the path probability by the probability assigned to $\mathbf{x}$ in the leaf.

As in [13], our implementation only uses events which are all the possible values of discrete variables (i.e., for every possible value we get one branch). This restriction simplifies the learning algorithm. It also means that the value of any variable appearing in the path is determined at the leaf; thus, at each leaf, we simply use a a joint distribution over the remaining variables. We have chosen a fairly simply representation at the leaves. The discrete variables are assumed to be independent, so we simply keep the marginal distribution of each one. (Dependency will be manifested as more splits in the tree.) The continuous variables are also independent of the discrete variables. The multivariate distribution over the discrete variables is represented as a mixture of Gaussians.

The basic operations which must be supported by density trees are computing the probability of some instantiation,

marginalization, introducing evidence into the tree (as the instantiation of the random variables in the tree) and sampling. All these operations are fairly easy to implement. Marginalization and instantiation can be done in time linear in the size of the tree, while sampling and finding the probability of some instantiation can be done in time linear in the depth of the tree. Note that other possible representations of a joint density (e.g., as a BN) do not necessarily have these linear time guarantees.

It remains only to describe the process by which the density tree, with the Gaussian mixture at the leaves, is constructed from our weighted samples. To build the tree, we start recursively from the root. At each node $u$, we determine which discrete variable we want to split on. As discussed in [13], the relative entropy between the empirical distribution in $u$ before and after the split is an appropriate error function. In the work of [13], uniform distributions were used at the leaf, leading to a very simple splitting criterion. In our case, uniform leaf distributions are clearly less appropriate; our representation of the leaf as a product of multinomials and a mixture of gaussians allows for much more accurate density learning. However, the associated splitting rule (using the relative entropy error function) is now much more expensive to compute. Instead, we used a very simple heuristic rule: splitting on the variable which partitions the samples as equally as possible among the branches. Even with this simple heuristic, the density trees yield good results. In order to avoid overfitting, we stop splitting a node when it contains less than some minimal number of samples.

Note that the splitting criterion and the splitting rule are based on the samples themselves, rather than on the weight. This property is particularly helpful in tailoring the estimate of the message to be accurate in areas where the clique potential is large. The samples are generated from the clique potential, so there will be more samples in regions where the clique potential has high density; this leads to a more refined partition in those regions, even if the message density there is low.

When the leaves are reached, the marginals of the discrete variables are estimated using a simple Bayesian estimation algorithm with a Dirichlet prior for regularization. The density over the continuous variables is estimated as a mixture of Gaussians, each with a diagonal covariance matrix. We execute this estimation using a variant of the standard EM algorithm [4] for Gaussian mixtures.

Unfortunately, EM is not as well-behaved for Gaussian mixtures as it is in discrete settings. The problem is that the likelihood of the data can be arbitrarily large, if we allow the variance of the estimated Gaussians to go to zero. There are several techniques for preventing this situation. We experimented with two of them. One is a simple and commonly used technique, where we assume that all of the

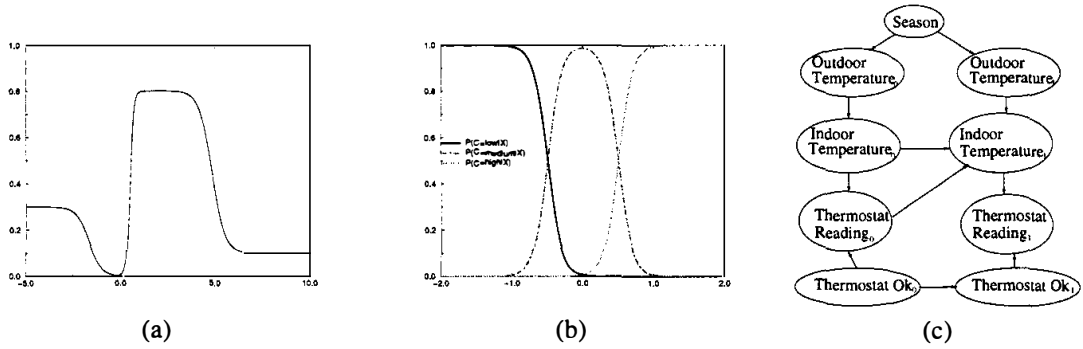

Figure 1: (a) and (b) Expressive power of a generalized softmax CPD. (c) The thermostat network.

Gaussians in the mixture have the same covariance matrix. The second, which turned out to work much better, is a version of EM that minimizes a regularized error function rather than the standard negative log-likelihood error function (this is analogous to the regularized error function for neural networks in [1]).

Specifically, let $\mathbf{y}[1], \ldots, \mathbf{y}[M]$ represent our training instances, each of which is an instantiation of the variables $Y_1, \ldots, Y_n$. (For simplicity, we ignore the weights on the instances.) Assume that we are trying to estimate the distribution as a mixture of $K$ Gaussians. The EM algorithm, in this setting, uses the current set of parameters $\theta$ (means and variances for the $K$ Gaussians and the weights of the mixture components) to estimate $P(k \mid \mathbf{y}[m], \theta)$ for every Gaussian $k = 1, \ldots, K$ and every data instance $m$. It then uses the results to minimize the negative log-likelihood function $-\ln \mathcal{L} = -\sum_m \ln P(\mathbf{y}[m])$ leading to the standard update rules for the means and covariances (see [1]). Instead, we try to minimize a regularized error function, which penalizes the algorithm for letting the variances get too small. Letting $\sigma_{ki}^2$ denote the variance of $Y_i$ in the $k$-th mixture component, we define our error function to be $-\ln \mathcal{L} + \lambda \sum_k \sum_{i=1}^n \frac{1}{2\sigma_{ki}^2}$, where $\lambda$ is a regularization coefficient that determines the extent of the penalty. Taking the derivative relative to $\sigma_{ki}$ and setting to zero, we obtain that the setting of $\sigma_{ki}$ that minimizes this function is

$$\sigma_{ki}^2 = \frac{\sum_m P(k \mid \mathbf{y}[m], \theta)(y_i[m] - \bar{y}_{ki})^2}{\sum_m P(k \mid \mathbf{y}[m], \theta)} + \frac{\lambda}{\sum_m P(k \mid \mathbf{y}[m], \theta)},$$

where $\bar{y}_{ki}$ is the mean of Gaussian $k$ in the $i$-th dimension. The first part of this expression is identical to the standard EM update rule for $\sigma_{ki}$. The second part causes the covariance to be larger than is warranted by the data. Note that the extent of the increase depends on the weight of samples assigned firmly to this mixture component: the more samples that are associated firmly with this Gaussian, the less effect the second term has on the variance. This is precisely the behavior we would like. Indeed, as we can see in the next section, the use of a regularization coefficient leads to significantly better results.

It is instructive to compare our algorithm to the simple pre-discretization approach, most often used in BNs. The main problem is that high-dimensional cliques over finely discretized variables are too large to fit in memory, making exact inference impractical. But even if we ignore the memory constraints, there are many domains where pre-discretization is impossible. For one, we may want to adapt the parameters as new data is obtained, requiring a re-discretization each time. Furthermore, in some cases, the function might be too expensive to evaluate more than a small number of times. One example of such a function comes up in the visual tracking domain, where we have a CPD defining the probability of an image given the state of the system. Computing this function involves a geometric projection and rendering, and is quite expensive.

## 5 Experimental Results

We tested our algorithm using two networks. The first is a continuous version of the real-life BAT network of [5]. In most of our experiments we used three time slices of the BAT network, leading to 72 variables of which 23 are continuous. We also wanted to test a network with continuous parents of a discrete node, which our version of BAT does not have. We constructed another small hybrid BN, shown in Figure 1(c), which models a simple thermostat. The *Thermostat Reading* variable is a classifier node, as described above.

To evaluate our algorithm we needed to find a good approximation to the ground truth. To do so, we discretized the continuous variables in the two BNs into 100 discrete values each. We then computed the KL-distance between the result of the exact inference in the discrete network and a discretization of the result of approximate inference in the hybrid network. We refer to this measure as the *KL-error*. Note that obtaining such an accurate discrete estimator cannot be done for general networks, as discussed before. In our experiments, we had to ensure that no clique in the tree contained more than four continuous variables.

We begin by investigating the influence of $\lambda$, the regular-ization coefficient used in EM. We first consider the pure density estimation problem, temporarily ignoring our algorithm. Figure 2(a) shows the result of fitting 10 Gaussians to a set of samples which was created during one of the runs (and is shown on the top right side) using three different values for $\lambda$. one of the runs using three different values for $\lambda$. In this example, $\lambda = 0.001$ overfits the data, $\lambda = 1000$ does not capture correctly the area in which the samples are concentrated, but $\lambda = 10$ seems to give a plausible result. Examining the KL-error after running our algorithm gives similar results. In this case, we averaged 5 runs over the thermostat network (querying the outside temprature at time 0, given the thermostat readings at both times). The graph shows the KL-errors after each iteration, where each iteration is defined as one pass of propagation in the clique tree (either up-stream or down-stream). When $\lambda$ is too small we get Gaussians with very small variance. As a result we get an estimation of almost 0 density in some areas where the true density is larger. This leads to a very big KL-error, as shown in the graph. On the other hand, when $\lambda$ is too big we get an estimation which is almost a uniform distribution. This gives a much smaller KL-error, but is not a useful estimation. Again, $\lambda = 10$ seems to give good results.

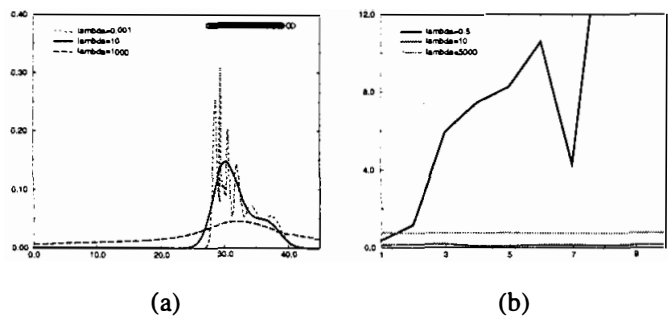

(a)                                    (b)

Figure 2: (a) Quality of density estimation as a function of $\lambda$. (b) KL-error as a function of $\lambda$.

The most important parameter which influences the running time of the algorithm is the number of samples drawn in every clique. The following table shows the time per iteration (on a Sun Ultra 2) on the BAT network as a function of the number of samples, and the average KL-error over 12 iterations. Here, there were 58 cliques in the join tree.

| samples | 100 | 200 | 500 | 1000 | 3000 | 5000 |
|---|---|---|---|---|---|---|
| iteration [sec] | 3.6 | 6.9 | 17.5 | 36 | 111 | 177 |
| avg KL-error | 0.562 | 0.254 | 0.233 | 0.151 | 0.051 | 0.066 |

The results are hardly surprising. The running time of the algorithm grows linearly with the number of samples and the error decreases when we increase the number of samples. A natural improvement to the algorithm is to start with a relatively small number of samples in order to get some initial estimate of the potentials, and then increase the number of samples in later iterations.

Like all sampling algorithms, our algorithm is sensitive to the likelihood of the evidence. However, we claim that the iterative nature of our algorithm solves this problem, at least partially. We tested our algorithm using the 3 time-slice BAT network. In every time slice, BAT contains the continuous variables $Xdot$ — the velocity of the car in the $X$ axis (i.e., left-right movement), and $Xdot$-sensed — the sensed value of $Xdot$. With probability 0.9999 the sensor is functional and the reading is $Xdot$ plus some gaussian noise. With probability 0.0001 the sensor is broken and the reading is uniform, independent of $Xdot$.

We tested two scenarios. In both of them we instantiated $Xdot$ at time 0 and $Xdot$-sensed at time 2 and queried $Xdot$-sensed at time 1. In the first case, we instantiated the variables to relatively close values, a likely situation. In the second, we used very different values, leading to a very unlikely scenario; in fact, using our exact discrete inference, we found that the probability of the sensor being broken in this case went up to 0.3. As expected, our algorithm performed much better in the first case. This is demonstrated in Figure 3. In (a) we see the density estimation of $Xdot$-sensed at time 1 for the easy case. As expected the algorithm converges quickly to the correct answer. In (b) we see the convergence in the second and more difficult case. This time the quality of the estimation is not as good and the convergence is slower. However, in some sense the second scenario is even more encouraging than the first. Our algorithm seems to slowly improve the quality of the estimation. The fact that the density does not go to zero far from the peak indicates that in later iterations the algorithm correctly identified the relatively high probability that the sensor is broken. Part (c) further demonstrates this point using the KL-error in the two cases.

Finally, it is interesting to compare the performance of our algorithm and Likelihood Weighting. To do a fair comparison, we gave each one of the algorithms the same CPU time and compared their KL-error. Since our algorithm takes advantage of the structure of the network to reduce to dimensionality of the samples drawn, we expected our algorithm to scale up much better when a lot of evidence is introduced to the network. To test this hypothesis we ran two experiments — in one we had only one evidence node while in the other we had 12 evidence nodes (note that we did not use unlikely evidence). In both cases we queried the same node ($Xdot$ at time 1). The results (averaged over ten runs) are given in Figure 4. In (a), we see that LW provides a rough initial estimate after a very short time, whereas our algorithm requires some time to perform the initial clique estimation and the first iteration. After this startup cost, it seems that our algorithm outperforms LW even for just one evidence node. The difference becomes much more dramatic in the presence of more evidence. While the performance of LW degrades severly, our algorithm is almost unaffected by the extra evidence.

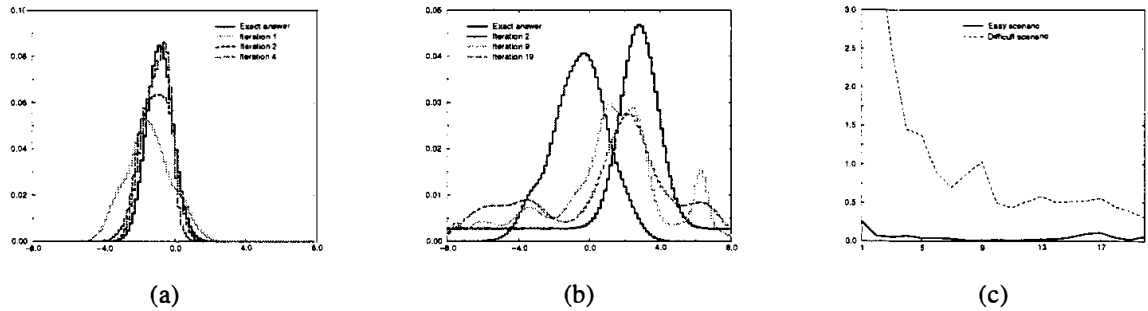

Figure 3: Effect of iterations: (a) Easy scenario. (b) Difficult scenario. (c) KL-error.

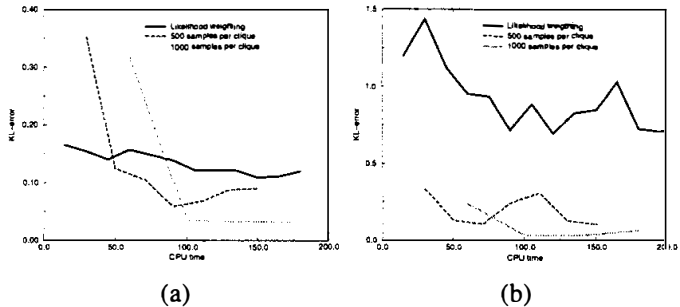

Figure 4: Comparison with LW: (a) One evidence node (b) 12 evidence nodes

## 6   Existing instances of our approach

Several known approximate inference algorithms turn out to be special cases of our general scheme. By viewing them from this perspective, we can understand their relationship to other inference algorithms. More importantly, we can sometimes point out ideas for possible improvements based on our general approach. We briefly sketch the crucial properties of the mapping.

**Conditional Gaussian (CG) networks [15]:** In this algorithm, the representation of the clique potentials is a table, with one entry for each combination of the discrete variables in the clique. In each entry, the table specifies a single multivariate Gaussian over the continuous variables in the clique. The approximate computation of messages is done using an "exact" algorithm, i.e., one that gets the best Gaussian approximation to the factor. Lauritzen shows that this algorithm preserves the correct first and second moments throughout the propagation, so that the algorithm is "exact" with respect to these. Thus, further iterations would not improve the quality of the result.

**Hybrid Propagation [3, 6] and** HUGS **[12]:** This family of algorithms is quite closely related to ours, as it also uses sampling as a substitute for exact operations within a clique. However, it is restricted to messages that are vectors of weighted samples, and does not use any type of density estimation for smoothing. More importantly, the algorithm only samples once in a clique, and the variable is then restricted to take on one of the sampled values. If the sampled values are poor (due, for example, to an uninformed sampling prior to message propagation), the resulting value space for the variable can potentially badly skew the remainder of the computation. None of these algorithms contain an iterative phase (such as ours); such a phase may help address this limitation.

**Dynamic discretization [14]:** This algorithm, which also works for hybrid BNs, is another instance of our general approach. It uses a piecewise-constant representation of the domain of the continuous variables in a clique. The representation is structured hierarchically, with regions split into finer pieces further down the tree. Unlike most of the other instances, this algorithm has an iterative phase. However, its iterations are limited to complete upward and downward passes. Our algorithm is more flexible in its ability to concentrate the computational efforts on some specific areas of the clique tree where the approximation is inaccurate.

**Likelihood Weighting [19]:** Roughly speaking, this algorithm samples from the BN, weighting each sample based on the likelihood of their evidence relative to this sample. It can be viewed as a special case of our algorithm where the clique tree has a single clique, containing all of the network variables. The representation of the densities is a set of samples, and only a single iteration is performed. If we introduce further iterations, the algorithm turns into a *bootstrap sampling* algorithm, where one set of samples is used as the basis for further sampling. If we add a density estimation step, the algorithm turns into a *smoothed bootstrap* algorithm.

**Survival of the Fittest (SOF) [11] and its extensions [13]:** These algorithm performs monitoring in a dynamic Bayesian network (DBN). It maintains a belief state — a distribution over the current state given the evidence so far. SOF maintains the belief state at time $t$ as a set of weighted samples. To get the time $t + 1$ samples, SOF samples from the time $t$ samples proportionately to their weight, stochastically propagates them to the next time slice, and weights them by the likelihood of the time $t + 1$ evidence. In terms of the general approach, the algorithm performs a single forward pass over the basic clique tree for the unrolled DBN: the one containing a clique for every pair of consecutive time slices. The message is generated using simple importance sampling. The extension of [13] can be viewed as augmenting the importance sampling at each clique with a density estimation phase.

# 7  Conclusions

In this paper, we presented a general approximate inference algorithm for Bayesian Networks. Our algorithm can best be viewed as a general schema which can be instantiated in different ways based on the representation scheme and the different ways to manipulate it. Many well known inference algorithms turn out to be a special case of this general view. Our approach combines the best features of exact and approximate inference: like approximate inference algorithms, it can deal with complex domains, including hybrid networks; like the clique tree algorithm, it exploits the locality structure of the Bayes net to reduce the dimensionality of the probability densities involved in the computation.

We described one particular instantiation of our algorithm that uses Monte Carlo Importance Sampling, combined with density estimation, in order to estimate the necessary functions. This particular instantiation is very general in the sense that all that is required from our factor representation is the ability to sample from factors. Hence, it allows us to provide a general-purpose approximate inference algorithm for arbitrary hybrid BNs. We have presented empirical results showing that our algorithm gives good results for nontrivial networks. We believe that our algorithm has the ability to scale up to significantly larger problems (such as visual tracking in complex environments), where existing inference algorithms are infeasible.

**Acknowledgments.**

We would like to thank Ron Parr, Xavier Boyen and Simon Tong for useful discussions. This research was supported by ARO under the MURI program, "Integrated Approach to Intelligent Systems," grant number DAAH04-96-1-0341, by DARPA contract DACA76-93-C-0025 under subcontract to IET, Inc., and by the generosity of the Powell Foundation and the Sloan Foundation.